\documentclass[conference]{IEEEtran}
\IEEEoverridecommandlockouts

\usepackage{cite}
\usepackage{amsmath,amssymb,amsfonts}
\usepackage{algorithmic}
\usepackage{graphicx}
\usepackage{textcomp}
\usepackage{xcolor}

\usepackage{booktabs}
\usepackage{multirow}
\usepackage{tikz}
\usetikzlibrary{patterns}
\usepackage{pgfplots}
\usepackage{filecontents}
\usepackage{pgfplotstable}
\def\BibTeX{{\rm B\kern-.05em{\sc i\kern-.025em b}\kern-.08em
    T\kern-.1667em\lower.7ex\hbox{E}\kern-.125emX}}
\begin{document}

% \title{Conference Paper Title*\\
% {\footnotesize \textsuperscript{*}Note: Sub-titles are not captured in Xplore and
% should not be used}
% \thanks{Identify applicable funding agency here. If none, delete this.}
% }

\title{Emotional Theory of Mind: Bridging Fast Visual Processing with Slow
Linguistic Reasoning\\
 % so delete this line
\thanks{This work was supported by NSERC Discovery Grant RGPIN/06908-2019.}

}
% {\footnotesize \textsuperscript{*}Note: Sub-titles are not captured in Xplore and should not be used}
% \thanks{Identify applicable funding agency here. If none, delete this.}
% }

% \author{Yasaman Etesam and \"{O}zge Nilay Yal\c{c}{\i}n and Chuxuan Zhang and Angelica Lim \\
%   Simon Fraser University, BC, Canada \\
%   \texttt{yetesam@sfu.ca,  oyalcin@sfu.ca, cza152@sfu.ca, angelica@sfu.ca } \\
\author{
\IEEEauthorblockN{Yasaman Etesam, \"{O}zge Nilay Yal\c{c}{\i}n, Chuxuan Zhang, Angelica Lim}
\IEEEauthorblockA{
\textit{Simon Fraser University, BC, Canada}\\
\{yetesam, oyalcin, cza152, angelica\}@sfu.ca}
}

% \author{Yasaman Etesam and \"{O}zge Nilay Yal\c{c}{\i}n and Chuxuan Zhang and Angelica Lim \\
%  Simon Fraser University, BC, Canada \\
%  \texttt{yetesam@sfu.ca,  oyalcin@sfu.ca, cza152@sfu.ca, angelica@sfu.ca }
%  }
% \author{\IEEEauthorblockN{1\textsuperscript{st} Given Name Surname}
% \IEEEauthorblockA{\textit{dept. name of organization (of Aff.)} \\
% \textit{name of organization (of Aff.)}\\
% City, Country \\
% email address or ORCID}
% \and
% \IEEEauthorblockN{2\textsuperscript{nd} Given Name Surname}
% \IEEEauthorblockA{\textit{dept. name of organization (of Aff.)} \\
% \textit{name of organization (of Aff.)}\\
% City, Country \\
% email address or ORCID}
% \and
% \IEEEauthorblockN{3\textsuperscript{rd} Given Name Surname}
% \IEEEauthorblockA{\textit{dept. name of organization (of Aff.)} \\
% \textit{name of organization (of Aff.)}\\
% City, Country \\
% email address or ORCID}
% }

\maketitle
\begin{abstract}
    The emotional theory of mind problem %in images is an emotion recognition task, specifically asking "How does the person in the bounding box feel?" 
requires facial expressions, body pose, contextual information and implicit commonsense knowledge to reason about the person's emotion and its causes, making it currently one of the most difficult problems in affective computing. In this work, we propose multiple methods to incorporate the emotional reasoning capabilities by constructing ``narrative captions" relevant to emotion perception, that includes contextual and physical signal descriptors that focuses on ``Who", ``What", ``Where" and ``How" questions related to the image and emotions of the individual. We propose two distinct ways to construct these captions using zero-shot classifiers (CLIP) and fine-tuning visual-language models (LLaVA) over human generated descriptors.
%using recent zero-shot classifier (CLIP), large vision language model (LLaVA) and large language models (GPT-4) on the Emotions in Context (EMOTIC) dataset. To evaluate a text-based language model on images, we construct "narrative captions" relevant to emotion perception, using a set of 891 physical social signal descriptions and 224 emotionally salient environmental context labels. 
We further utilize these captions to guide the reasoning of language (GPT-4) and vision-language models (LLaVa, GPT-Vision). We evaluate the use of the resulting models in an image-to-language-to-emotion task. Our experiments showed that combining the ``Fast" narrative descriptors and ``Slow" reasoning of language models is a promising way to achieve emotional theory of mind.

\end{abstract}
% Add something about implications

%posture,mouth,laughing,nose,eye, gaze, hand arm, head,eyebrow,neck,finger,leg,touch,body,speech,leaning,walking,other (18)
% try to order it in decreasing order and with different number of social signals
% signing a football
% do clip on the person cut-out only
% do clip on the entire image for location
% do clip 
% \end{abstract}


\section{Introduction}
Imagine a photo depicting a mother looking up from a swimming pool at her 4-year old daughter perched at the edge of a high diving board. How does this person feel? Traditional computer vision approaches might detect the swimming activity and infer a positive emotion. However, considering her perspective on the invisible consequences of the diving activity, including the possibility of her daughter falling off the diving board, a more plausible emotion would be fear. 
Humans can adeptly combine fast visual processing (e.g. detecting the swimming activity) and slow reasoning (e.g. possibility of falling off) capabilities~\cite{daniel2017thinking} to perform emotional theory of mind.

% Humans adeptly combine fast visual processing and slow reasoning capabilities~\cite{daniel2017thinking} to perform emotional theory of mind. 

%This phenomenon is also known under the concepts of affective and cognitive empathy~\cite{reniers2011qcae}, which is akin to the difference between mimicking one's emotional expression versus understanding the causes of the emotion for perspective taking and targeted helping. 

Indeed, our ability to recognize emotions and their causes allows us to understand one another,  foster enduring social bonds, and engage in socially acceptable interactions. Integrating emotion recognition capabilities into our technologies holds promise for enhancing and streamlining human-machine interactions  \cite{picard2000affective}. Emotion recognition in humans requires much more than detecting patterns. Understanding of the causal relations, contextual information, social relationships as well as using theory of mind, all contribute to the complexity of this task, which are unresolved problems in affective computing research. However, emotion recognition systems today still suffer from poor performance \cite{barrett2019emotional} due to the complexity of the task. Many image-based emotion recognition systems focus solely on using facial or body features \cite{pantic2000expert,schindler2008recognizing}, which can have low accuracy in the absence of contextual information \cite{barrett2011context,barrett2017emotions}.

% The affective computing research community has been moving towards creating datasets and building models that include or make use of contextual information that is required for emotional reasoning and theory of mind tasks. 
Due to the importance of contextual information for emotional reasoning and theory of mind tasks, the affective computing research community has created datasets and built models that include or make use of context.
As a recent example, the EMOTIC dataset incorporates contextual and environmental factors for emotion recognition in still images~\cite{kosti2019context}. The inclusion of contextual information in addition to facial features is found to significantly improve the accuracy of the emotion recognition in a multi-labelling task \cite{le2022global,mittal2020emoticon,wang2022context}. 
% However, using this information to infer the emotions of others has not been examined, as it requires commonsense knowledge and high-level cognitive capabilities such as reasoning and theory of mind which are unsolved issues in affective computing~\cite{ong2019computational}. 

% Large Language Models (LLMs) that are based on the Transformer architecture \cite{vaswani2017attention} have been shown to excel at Natural Language Processing (NLP) tasks \cite{brown2020language, chowdhery2022palm}, offering a way to achieve emotional theory of mind through linguistic descriptors. LLMs gained success in increasing accuracy and efficiency in NLP problems including multimodal tasks such as Visual Question Answering \cite{antol2015vqa} and Caption Generation \cite{vinyals2015show}. Recently, they have been also used in commonsense reasoning \cite{sap2019socialiqa,bisk2020piqa,li2022systematic}, emotional inference \cite{mao2022biases} and theory of mind \cite{sap2022neural} tasks, however their capabilities on emotional theory of mind in visual emotion recognition tasks have not been explored.

% Until now, images of humans experiencing emotions have been analyzed based on immediately visible characteristics (e.g. face, body, activity and physical relationships) ~\cite{kosti2019context}, with results that struggle to access second order thinking of non-visible consequences. 
\begin{figure}[t]
\centering
\includegraphics[width=0.98\linewidth]{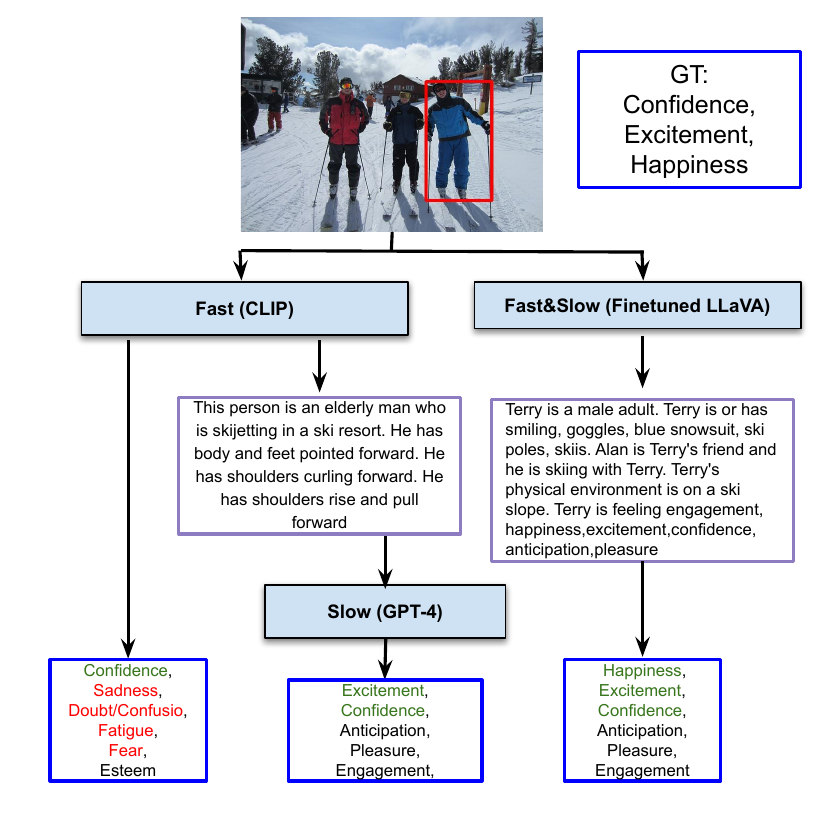}
\vspace{-4mm}
\caption{
Fast frameworks might perceive the body pose with raised arms and predict the emotions of `surprise' and `fear'. Slow networks could further reason about context and future implications.}
\label{fig:teaser}
\vspace{-4mm}
\end{figure}

In this paper, we examine different approaches to incorporate the ``Slow" reasoning processes into the ``Fast" emotion recognition pipelines. We
focus on a multi-label, contextual emotional theory of mind task by creating a pipeline that includes caption generation and LLMs that combines ``Fast" and ``Slow" processing (see Fig.~\ref{fig:teaser}). Towards visual grounding of text, we use a zero-shot classifier (CLIP) to generate explainable linguistic descriptions of a target person in images using physical signals, action descriptors and location information. We then use an LLM (e.g. GPT-4) to reason about the generated narrative text and predict the possible emotions that person might be feeling. We further use a chain-of-thought method on vision-language models (i.e., LLAVA~\cite{liu2023visual}) to guide their reasoning capabilities.
We compared our results with one-shot (i.e., fast processing) vision models and vision language models, which can combine both fast visual processing and slower, cognitive and linguistic processing.
% Here we need to add all the steps we took and contributions.
% Our results showed????

The contributions of this paper are as follows:
    %\vspace{-2mm}
\begin{itemize}
    \item Comparing the “Fast” (CLIP) visual framework to the ``Fast and Slow" visual and language reasoning frameworks (NarraCaps + LLM, ExpansionNet + LLM, NarraCapsXL) for contextual emotional understanding.
    \item Introducing two ways of achieving "Fast \& Slow" Framework through generating zero-shot and fine-tuned narrative descriptors of the images.
    %\item Generating fine-tuned \emph{narrative captions} using LLaVA (NarraCapsXL), an interpretable textual representation for images of people experiencing emotions
    % \item Generating fine-tuned \emph{narrative captions} (NarraCapsXL), an interpretable textual representation for images of people experiencing emotions.
    %\vspace{-2mm}
   % \item Investigate the emotion inference capabilities of large vision language models and large language models (here, CLIP, GPT-3.5, MPT, Falcon, Baichuan, LLaVA);
    %\vspace{-2mm}
    \item Investigating the use of ``Chain-of-thought" prompting methods to mimic emotional reasoning steps in large vision language models (LLaVA). 
\end{itemize}

\section{Related Work}
Cognitive processes can be categorized into two groups, referred to as intuition and reason, or System 1 and System 2 \cite{daniel2017thinking}. In this dual-process model, one system rapidly generates intuitive answers to problems as they arise, while the other system monitors the quality of these proposals using reasoning ~\cite{lieberman2007social, evans2008dual,kahneman2002representativeness}. %Regarding empathy, a general term encompassing various neurocognitive functions, the data suggest that a two-factor structure (affective and cognitive empathy) provides the best and most parsimonious fit \cite{reniers2011qcae, blair2005responding}.
% https://docs.google.com/document/d/1rl2GsR-zDFeEPVz6HhgrI6f9MJYquTZdLEw_3D5qgrs/edit#
Thinking Fast (System 1) operates quickly and intuitively, based on heuristics and shortcuts, and is the kind of thinking we use for everyday decisions. This process resembles traditional recognition and categorization models. In comparison, Thinking Slow (System 2) involves more logical and effortful thinking. This system is engaged when we face complex problems that require reasoning, causal analysis and decision making. These processes are suggested to be related to linguistic reasoning processes in humans. Human visual emotion recognition, similar to many complex tasks, rely on both fast judgements related to facial and bodily expression detection which are automated and intuitive (System 1),  as well as understanding contextual and causal relationships between the actors and their surroundings (System 2). 

Current emotion recognition models suffer from relying only on fast recognition models that resemble System 1, which recently ignited discussions in the Affective Computing community on the importance of context and reasoning \cite{barrett2019emotional}. 
Most of the prior work on visual emotion recognition tasks (see \cite{ko2018brief, mellouk2020facial} for recent reviews) focus on facial expressions \cite{pantic2000expert} or body posture \cite{schindler2008recognizing} to predict emotions in usually a few number of emotion categories such as happy, sad, angry \cite{ekman1971constants}. However, facial and posture information alone would not be sufficient to understand the emotional state of a person, as humans rely on contextual information to reason about the social and environmental causes of one's emotion \cite{barrett2011context}. 
There are recent attempts to incorporate contextual information to visual emotion recognition datasets and models (i.e., EMOTIC~\cite{kosti2019context}. 
%A number of computer vision approaches have been developed in response to the release of the EMOTIC dataset. 
The EMOTIC baseline uses one CNN to 
%extracts features from the target human, as well as the entire image.
 extract features from the target human, and another to obtain global features over the entire image, and incorporates the two sources using a fusion network ~\cite{kosti2019context}. %This resulted in a mean average precision (mAP) of $28.33$ over all 26 categories. 
% Subsequent fusion methods incorporated body and context visual information \cite{huang2021emotion} at global or local scales \cite{le2022global}, investigated contextual videos \cite{lee2019context}, or worked to improve subsets of the EMOTIC dataset, such as \cite{thuseethan2021boosting} which considered the photos only including two people. 
In PERI \cite{mittel2023peri}, the modulation of attention at various levels within their feature extraction network using feature masks.
% , resulted in an mAP of 33.33. 
To the extent of our knowledge, the current best approach was Emoticon proposed by Mittal and colleagues \cite{mittal2020emoticon}, which explored the use of graph convolutional networks and image depth maps, they found that using depth maps could improve the results.
% the mAP to $35.48$. 
%Overall, the latest results leave room for improvement.
However, the emotional theory of mind and reasoning mechanisms are poorly understood within the affective science and social psychology community \cite{ong2019computational} and has not been examined or incorporated in contextual emotion recognition models.

\emph{Natural language and emotional theory of mind.} Language is a fundamental element in reasoning and it plays a crucial role in emotion perception~\cite{lindquist2015role, lieberman2007putting, lindquist2006language, lindquist2013s, gendron2012emotion}. 
% Here importance of language in reasoning and emotional theory of mind ... "Slow" reasoning capabilities & ways in which to incorporate slow language reasoning to 
% In NLP, significant work has focused on automatic sentiment detection that focuses on text polarity (positive or negative); a review is available in \cite{nandwani2021review}, and is out of the scope of this paper. Instead, 
%In this work, we focus on the body of work in English literature that discusses how writers use empathy to describe characters, their actions and external cues, and ``seek to evoke feelings in readers employing the powers of narrativity." \cite{keen2015narrative}. Writing textbooks such as The Emotion Thesaurus \cite{puglisi2019emotion} provide suggestions for sets of visual cues or actions that support emotional evocation, as a guide for allowing readers to imagine the most relevant features of the person in the scene. For example, to evoke curiosity% (see Table~\ref{tab:physical-signals})
%, a writer may narrate, "she tilted the head to the side, leaning forward, eyebrows furrowing" or "she raised her eyebrows and her body posture perked up." To the best of our knowledge, a gap in the emotional theory of mind task remains in testing how large language models (LLMs) and language-vision models (LVMs) could be used to describe and reason upon a visual snapshot of a person experiencing an emotion at a given moment in time, rather than reasoning about a sequence of events.
%
%\emph{Large language models and theory of mind.} 
Recent investigations into large language models (LLMs) have uncovered some latent capabilities for reasoning \cite{huang2023towards}, including some sub-tasks on emotion inference \cite{mao2022biases}. Emotional theory of mind tasks using language tend to focus on appraisal based reasoning on emotions, inferring based on a sequence of events. For instance, among other social intelligence tasks, further research \cite{sap2022neural} explored how a language model could respond to an emotional event, e.g. ``Although Taylor was older and stronger, they lost to Alex in the wrestling match. How does Alex feel?" Their findings suggested some level of social and emotional intelligence in LLMs.
% , finding the best performance with GPT-3-davinci. While emotion and mental state inference remains an elusive goal for AI systems \cite{choi2022curious}, this compelling result suggests an avenue for exploring the contextual emotion recognition problem using natural language as an intermediary. 
Further works used chain-of-thought prompting methods on complex reasoning tasks, by
generating intermediate guiding prompts to assist the reasoning process \cite{wei2022chain}. Incorporating explicit linguistic knowledge through these type of prompting mechanisms is suggested to be a promising way to assist the reasoning and implicit knowledge in LLMs \cite{manning2022human, qiao2023reasoning}. 

\emph{Vision language models.} One natural approach to link the visual emotional theory of mind task to the capabilities of large language models is to explore the large body of work in image captioning \cite{stefanini2022show} and visual question and answering \cite{zhu2016visual7w}. Early work related to emotional captions included Senticap \cite{mathews2016senticap} and Semstyle \cite{Mathews_2018_CVPR} which focused on the emotional style of the resulting caption, rather than an emotional theory of mind task. For example, a caption with sentiment generated by Senticap changed the neutral phrase, ``A motorcycle parked behind a truck on a green field." to ``A beat up, rusty motorcycle on unmowed grass by a truck and trailer." More recent self-supervised vision language models such as Visual-GPT \cite{Chen_2022_CVPR} for image captioning and CLIP \cite{radford2021learning,cornia2021universal} provide zero-shot captioning \cite{barraco2022unreasonable} on datasets of images and their captions of everyday activities and objects. Amidst the success of visual language models in tasks like image classification~\cite{pratt2023does, shu2022test, qiu2021vt} and object detection~\cite{gu2021open, kuo2022f, kim2023region, long2023fine}, the opportunity still remains to generate captions that describe details relevant to understanding the emotional state of the person in the image.

\section{Methodology}

% TO PUT BACK IN FINAL PAPER
% \input{tables/qualitative}
% \input{tables/table1_pdf}
%\input{tables/table2_pdf}

% \input{tables/qualitative_failure}
% \yasaman{In this paper using model families offered by OpenAI\footnote{https://openai.com/} we compare, Fast and Slow zero shot frameworks with a Fast method(CLIP~\cite{radford2021learning})}

In this paper, we propose and evaluate ways to model Emotional Theory of Mind by combining the ``Fast" visual processing capabilities using visual classification models and the ``Slow" reasoning capabilities using language models. 
%we aim to compare traditional "Fast", vision-only classification methods with a newly proposed "Fast and Slow" framework for emotion inference.
% another way to position ourselves would be including emotional reasoning to emotion recognition, without focusing on fast-slow processing... 
%
%\subsection{Emotion Inference: Fast and Slow}
%We propose a "Fast and Slow" emotion inference that first provides 
The ``Fast" System 1 provides visual information related to human affect that can be directly captured from the images, e.g. ``His veins were popping out from his skin. His cheeks were red and eyebrows narrowed." We then employ language models to provide cognitive reasoning on the observation, akin to System 2 ``Slow" processing, e.g. ``Adam is running in a marathon." 

To combine these two processing capabilities, we create linguistic descriptors of the images (Fast processing), to guide the reasoning processes (Slow processing) in Large Language or Vision-Language models (see Fig. \ref{fig:teaser}), similar to ``Chain-of-Thought" prompting methods \cite{wei2022chain}. We examine two different ways in constructing these linguistic descriptors, which can be considered as image captions: 1) Zero-shot NarraCaps and 2) Fine-tuning VLMs using Human-generated captions NarraCapsXL.
%To do this, we first generate a caption of the image, then use an LLM for implicit cognitive reasoning during emotion inference (see Fig. \ref{fig:teaser}). 
\subsection{Narrative Captioning}
% Our first captioning method, which we call Narrative Captioning (NarraCaps),
The first captioning method which is called Narrative Captioning (NarraCap)~\cite{etesam2024contextual} makes use of templates and the vision language model CLIP~\cite{radford2021learning}. We also test a baseline captioning method, ExpansionNet \cite{hu2022expansionnet}, and fine-tune a vision-language model LLaVA \cite{liu2023visual} that was not specifically trained for emotion understanding.
% We build upon previous captioning works that describe a person and their activity.

For generating NarraCaps, following previous work ~\cite{etesam2024contextual}, we extract the cropped bounding box of a person from an image and provide gender/age information to CLIP for identification. Then, we analyze the entire image to understand the depicted actions using a list of 848 actions sourced from Kinetics-700 \cite{smaira2020short}, UCF-101 \cite{soomro2012ucf101}, and HMDB datasets \cite{kuehne2011hmdb}.
\textit{Narrative captions} enrich the image description by incorporating over 850 social signals gathered from a writer's guide that provides suggestions for sets of visual cues or actions that support emotional evocation \cite{puglisi2019emotion}. These signals, along with the cropped bounding box, are fed to CLIP to generate descriptive captions for the person in the image. To provide further context, we utilize 224 environmental descriptors from guides to urban \cite{puglisi2016urban} and rural \cite{puglisi2016rural} settings to describe the scene's location. Examples of narrative captions (NarraCaps) can be found in Fig.~\ref{tab:images}, as well as failure cases in Fig.~\ref{tab:fail}.

\subsection{Human-generated Captions}
\label{human-generated-cap}
% We then produce our results with human captioning on a small sample, both with and without fine-tuning vision-language models.
We then employ fine-tuning and a small set of human-generated captions to generate human-like captions for images.
To do so, we use emotional descriptions generated by human participants for 387 samples of the EMOTIC database \cite{yang2023contextual}. The descriptions include the age, gender, profession, physical signals, human interactions, and environmental features for each person. Age and gender information is captured via using one of the categories among ``adult", ``child" or ``adolescent", and name of the person (e.g., ``Sean" to indicate male, ``Jane" for female). 
% We use these captions to fine-tune the open-source Vision-Language Model (LLaVA), and generate similar captions given each image. We refer to this fine-tuned captioning method as NarraCapsXL. 
% % Some details on this process
% \subsection{Finetuning VLM}

Using LoRA~\cite{hu2021lora}, we fine-tuned LLaVA\footnote{https://github.com/haotian-liu/LLaVA} on these human-generated captions~\cite{yang2023contextual} and ground truth labels from EMOTIC~\cite{kosti2019context}. Since human-generated captions are available for only a limited set of images, we increased the dataset size by augmenting the data and using each image ten times with random permutations of labels.
The prompt we used for finetuning and inference is:
\\
\textit{``Write a caption describing the person in the red bounding box. Then from suffering, pain, aversion, disapproval, anger, fear, annoyance, fatigue, disquietment, doubt/confusion, embarrassment, disconnection, affection, confidence, engagement, happiness, peace, pleasure, esteem, excitement, anticipation, yearning, sensitivity, surprise, sadness, and sympathy, pick the top labels that this person is feeling at the same time."}

And for the answer format for fine-tuning, we experimented with both  \textit{$<$caption$>$ + This person is feeling $<$labels$>$} and \textit{$<$caption$>$ + $<$name of the target\> is feeling $<$labels$>$}. Note that the name of the target is the first word in the human captions. 
Using this fine-tuned model, we then generate similar captions given each image. We refer to this fine-tuned captioning method as NarraCapsXL (NarraCaps X LLaVA). 

\subsection{Emotion Inference: Fast and Slow}
\label{sec:inference}
Using the captions generated either using the zero-shot NarraCaps method, or the fine-tuned NarraCapsXL method, as described above, we incorporate explicit linguistic knowledge to the Large Language Models. By doing so, we guide the reasoning process of LLMs, similar "Chain-of-thought" prompting methods used in prior work \cite{wei2022chain, manning2022human, qiao2023reasoning }. 

% Maybe a figure showing different captions vs. CLIP vs. language models.

The generated caption is sent to the LLM to obtain a set of emotions that is experienced by the individual in the bounding box. To do this, we provide the caption, along with a prompt, to GPT-4 with the temperature set to zero. The prompt asks for the top emotion labels understood from the narrative caption: %(see \ref{sec:appendix}.2)%, denoted by <caption>
\\
``\textit{$<$caption$>$ From suffering, pain,}
\textit{[...], and sympathy, pick the top labels that this person is feeling at the same time.}" \label{gpt-prompt}

% Six labels were requested from GPT as it is the average number of ground truth labels in the validation set and we use this number in CLIP, random, and majoroty baselines. We also used the label descriptions from EMOTIC to be consistent with the annotators.

%For generalization and reproducibility purposes, we also tested our narrative captions with three open-source Large Language Models: Baichuan-chat-13B\footnote{https://github.com/baichuan-inc/Baichuan-13B}, Falcon-Instruct-7B~\cite{almazrouei2023falcon}, and MPT-Chat-7B~\cite{mosaicml2023introducing}. However, due to their significantly lower number of parameters compared to GPT, they did not perform as well. These models achieved mAP scores of $17.12$, $17.10$, and $17.11$, respectively, which are considerably lower than GPT-3.5's score of $18.58$ and GPT-4's score of $19.22$.

\section{Experiments}

% \begin{table*}[htbp]
% \centering
% \caption{Performance Metrics of Various Models - Micro average}
% \label{my-label}
% \begin{tabular}{@{}lS[table-format=1.2]S[table-format=1.2]S[table-format=1.2]S[table-format=1.2]S[table-format=1.2]S[table-format=1.2]@{}}
% \toprule
% {Model} & {Precision (\%)} & {Recall (\%)} & {F1 Score (\%)} & {Hamming Loss (\%)} & {Jaccard Index (\%)} & {Subset Accuracy (\%)} \\ \midrule
% Emotic            & 43.56 & 61.04 & 50.84 & 20.03 & 34.08 & \textbf{2.57} \\
% Majority          & 49.45 & 67.20 & \textbf{56.98} & 17.24 & \textbf{39.84} & 0.76 \\
% Rand6             & 16.84 & 22.89 & 19.40 & 32.29 & 10.74 & 0.00 \\
% Rand6-weighted    & 37.69 & 51.21 & 43.42 & 22.67 & 27.73 & 0.01 \\
% Clip              & 18.06 & 24.54 & 20.81 & 31.72 & 11.61 & 0.00 \\
% Action            & 40.85 & 55.40 & 47.03 & 21.20 & 30.74 & 0.45 \\
% Expnet            & 36.33 & 49.35 & 41.85 & 23.29 & 26.46 & 0.48 \\
% Narracap + gpt3.5 & 35.05 & 47.56 & 40.36 & 23.87 & 25.28 & 0.25 \\
% Narracap + gpt4   & 43.40 & 58.62 & 49.88 & 20.01 & 33.23 & 0.29 \\
% LLaVA             & \textbf{54.14} & 36.47 & 43.58 & 16.04 & 27.86 & 0.78 \\
% GPT4 vision       & 50.09 & 61.96 & 55.40 & 16.95 & 38.31 & 0.67 \\
% finetuned LLaVA   & 31.31 & \textbf{77.91} & 44.67 & \textbf{13.18} & 28.76 & 1.53 \\
% \bottomrule
% \label{table-baselines}
% \end{tabular}
% \end{table*}

\begin{table*}[h!]
\centering
\caption{Performance Metrics of Various Models-macro average on EMOTIC Test Set}
\begin{tabular}{
  lllccc
}
\toprule
\textbf{Framework} & \textbf{Method} & \textbf{Model} & \textbf{Precision (\%)} & \textbf{Recall (\%)} & \textbf{F1 Score (\%)}  \\
\midrule
%Emotic            & 25.02 & 35.07 & 28.83 & 19.35 & 19.24 & \textbf{2.73} \\
Baseline & Trained & EMOTIC            & $25.02^{\pm 0.28}$ & $35.07^{\pm 0.49}$ & $28.83^{\pm 0.33}$  \\
%Majority          & 11.41 & 23.08 & 15.01 & 17.24 & 11.41 & 0.76 \\
%Majority & $11.41^{\pm 0.05}$ & $23.08^{\pm 0.00}$ & $15.01^{\pm 0.05}$ & $17.24^{\pm 0.09}$ &  $0.76^{\pm 0.09}$ \\
%Rand6             & 16.90 & 23.12 & 14.90 & 32.29 & 8.37  & 0.00 \\
%Rand6             & $16.90^{\pm 0.13}$ & $23.12^{\pm 0.34}$ & $14.90^{\pm  0.15}$ & ${32.29}^{\pm 0.07}$ & $0.00^{\pm 0.00}$ \\
%Rand6-W    & 17.02 & 23.17 & 19.45 & 22.67 & 12.81 & 0.01 \\
%Rand6-weighted    & $17.02^{\pm 0.15}$ & $23.17^{\pm 0.20}$ & $19.45^{\pm 0.17}$ & $22.67^{\pm  0.08}$ & $0.01^{\pm 0.01}$ \\
\midrule
%CLIP              & 21.77 & 28.58 & 16.97 & 31.72 & 9.70  & 0.00 \\
Fast & Zero-shot & CLIP          & $21.77^{\pm 0.19}$ & $28.58^{\pm 0.35}$ & $16.97^{\pm 0.18}$  \\
%Action + GPT3.5   & 20.34 & 27.45 & 21.19 & 21.20 & 0.45 \\
%Expnet + GPT3.5   & 21.51 & 30.76 & 21.83 & 23.29 & 14.18 & 0.48 \\
%NarraCaps+GPT3.5   & 21.35 & 30.42 & 22.51 & 23.87  & 0.25 \\
%NarraCaps+GPT4     & 25.98 & 31.25 & 26.11 & 20.01 & 17.23 & 0.29 \\
Fast\&Slow & Zero-shot & ExpNet + GPT4          & $24.94^{\pm 0.58}$ & $23.57^{\pm 0.27}$ & $22.29^{\pm 0.30}$  \\

Fast\&Slow & Zero-shot & NarraCaps + GPT4     & $25.50^{\pm 0.32}$ & $33.37^{\pm 0.42}$ & $26.67^{\pm 0.30}$  \\

Fast\&Slow & Fine-tuned &  NarraCapsXL + LLaVA     & $28.83^{\pm 0.28}$ & $\textbf{40.51}^{\pm 0.41}$ & $\textbf{31.05}^{\pm 0.29}$ \\

Fast\&Slow & Fine-tuned & NarraCapsXL + GPT4  & $28.78^{\pm 0.31}$ & $37.67^{\pm 0.42}$ & $30.19^{\pm 0.31}$ \\

\midrule

VLM  & Zero-shot & LLaVA           & $\textbf{33.78}^{\pm 0.86}$ & $21.38^{\pm 0.30}$ & $22.86^{\pm 0.32}$ \\
%LLaVA             & 27.77 & 18.51 & 19.58 & 16.04 & 12.72 & 0.78 \\
%Gpt4 vision       & 37.48 & \textbf{38.35} & \textbf{34.47} & 16.95 & \textbf{23.05} & 0.67 \\
VLM & Zero-shot & GPT4-vision       & $29.07^{\pm 0.37}$ & $27.48^{\pm 0.37}$ & $26.12^{\pm 0.28}$  \\
VLM & Fine-tuned & LLaVA  & $28.86^{\pm 0.30}$ & $35.67^{\pm 0.42}$& $28.27^{\pm 0.30}$ \\
%Finetuned LLaVA   & \textbf{53.08} & 16.01 & 21.77 & \textbf{13.17} & 14.19 & 1.53 \\
\bottomrule
\label{table-baselines}

\vspace{-4mm}
\end{tabular}
\end{table*}

\begin{figure*}[ht!]
% \vspace{-15mm}
\centering
\includegraphics[width=0.98\linewidth]{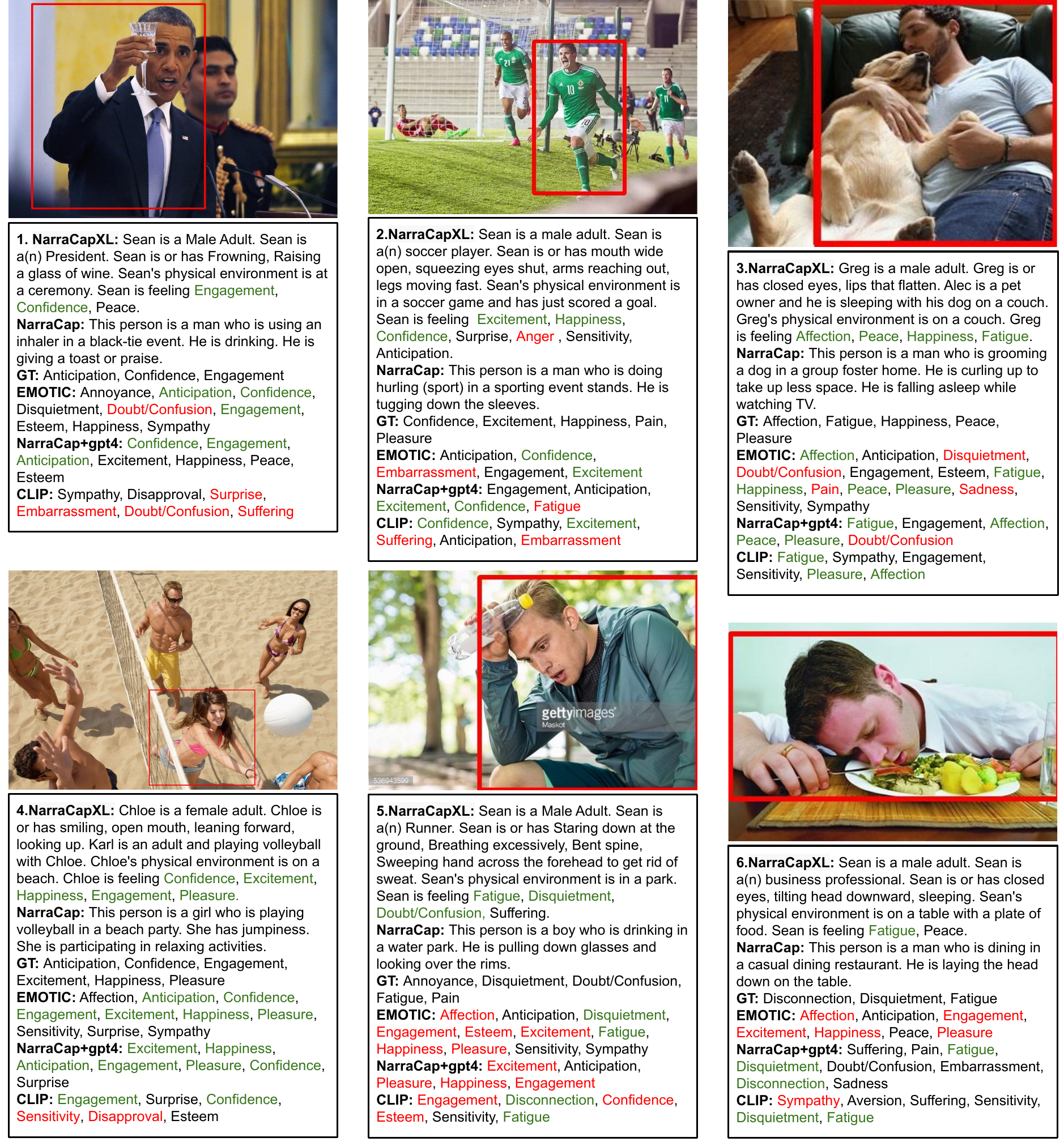}
%\framebox(500,300){}
%\framebox[\textwidth]{\rule{0pt}{140pt}}
% \vspace{-2mm}
 \caption{Qualitative results of EMOTIC images, ground truth (GT) labels, captions and inferred labels from example models from "Fast" and "Fast \& Slow" framework.}
    \label{tab:images}
\vspace{-4mm}
\end{figure*}
% }]

\begin{figure*}[t]
\includegraphics[width=\textwidth]{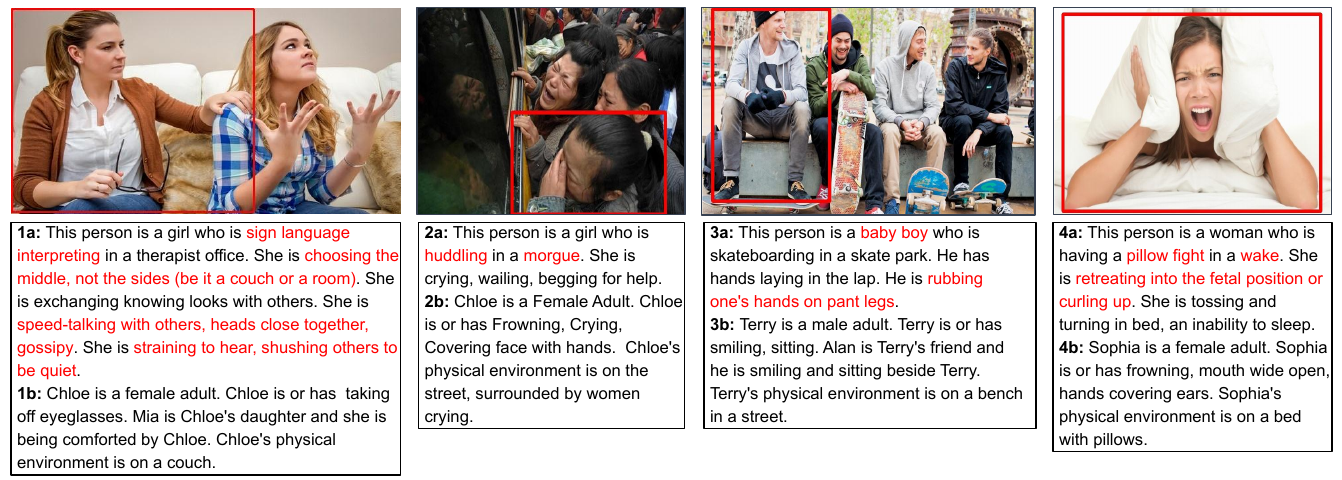}
\vspace{-6mm}
\caption{Example failure cases of NarraCap (X.a) that was correctly captioned by trained LLaVA captions (X.b).}
\label{tab:fail}
\vspace{-4mm}
\centering
\end{figure*}

We use the EMOTIC dataset which covers 26 different labels. The related emotion recognition task is to provide a list of emotion labels that matches those chosen by annotators.
% , responding to the question of, ``How does the person in the bounding box feel?". 
The training set (70\%) was annotated by 1 annotator, where Validation (10\%) and Test (20\%) sets were annotated by 5 and 3 annotators, respectively. For evaluation metrics we used precision, recall, F1 score,
implemented using the scikit-learn library \cite{scikit-learn}. As our methods output labels rather than probabilities, we cannot utilize mAP as an evaluation metric. We %limit our scope to zero-shot methods across the 
use varying state-of-the-art model families offered by OpenAI\footnote{https://openai.com/}, including a zero-shot classifier (CLIP), a large language model (GPT-4, hereafter also referred to as GPT), and an large vision language model (GPT4-vision). We also use an open-source VLM (LLaVA~\cite{liu2023visual}). We compare the following methods. %LLAVA fine-tuning was performed on four A40 48GB GPUs.

\subsection{Trained Baseline Model}
\textbf{EMOTIC} Along with the dataset, Kosti and colleagues \cite{kosti2019context} introduced a two-branched CNN-based network baseline. %The first branch is a feature extraction module which gets the bounding box of the target person as input, and the second branch is an image feature extraction which gets the whole image as the input. 
The first branch extracts body related features and the second branch extracts scene-context features. Then a fusion network combines these features and estimates the output, which is resulted in a mAP of 28.33. Further work such as \cite{mittal2020emoticon} using pose and face features (context1), background information (context2), and interactions/social dynamics (context3), reported higher mAP results (mAP = 35.48, F1 not reported), however, could not be included in our analysis as we were not able to reproduce their results.
% \yasaman{should we specify which loss? emoticon hasn't}
%\textbf{Emoticon} Motivated by Frege's principle \cite{resnik1967context}, Mittal and colleagues \cite{mittal2020emoticon} proposed an approach by combining three different interpretations of context. They used pose and face features (context1), background information (context2), and interactions/social dynamics (context3). They used a depth map to model the social interactions in the images. Later, they concatenate these different features and pass it to fusion model to generate outputs. \cite{mittal2020emoticon}
% \yasaman{shold we specify that we unconscioussed the depth based one?}

%\subsection{Random and Majority}

%\textbf{Majority} This Majority baseline selects the top $6$ most common emotions in the validation set as the predicted labels for all test images (\textit{Majority}). 

%\textbf{Random} We consider selecting either $6$ emotions randomly from all possible labels (\textit{Rand6}) or selecting $6$ labels randomly where the weights are determined by the number of times each emotion is repeated in the validation set (\textit{Rand6-W}). 

\subsection{Fast Framework}
\textbf{CLIP-only} We evaluate the capabilities of the zero-shot classifier CLIP to predict the emotion labels. In this study we pass the image with the emotion labels in the format of: \textit{``The person in the red bounding box is feeling \{emotion label\}"} to produce the probabilities that CLIP gives to each of these sentences. We then select the $6$ labels (average number of ground truth labels in validation set) with the highest probabilities.

\subsection{Fast and Slow Framework}
\label{sec:fastslow}
%\textbf{Action and LLM} We consider only the \textbf{what} portion of the caption and  disregard physical signals, environment and age. We pass the set of actions to CLIP and generate a single action caption per image in the form of \textit{This person is [activity]}. We then send this caption to GPT-3.5 and obtain 6 possible emotions using the same prompt template as described in Sec. 3.3. 

\textbf{ExpansionNet (ExpNet) and LLM} ExpansionNet v2 \cite{hu2022expansionnet} is a fast end-to-end training model for Image Captioning. Its characteristic is expanding the input sequence into another length with a different length and retrieving the original length back again. The model achieves the state of art performance over the MS-COCO 2014 captioning challenge, and was used as a backbone for a recent approach trained on EMOTIC~\cite{de_Lima_Costa_2023_CVPR}.
We fed the captions generated by ExpansionNet v2 into GPT-4 to obtain top emotion labels.

\textbf{Narrative captioning (NarraCaps) and LLM}
Using narrative captions, we pass them to Large Language Models and ask for the related emotion labels using the prompt described in Sec.~\ref{gpt-prompt} For LLMs, we used GPT-4 with temperature parameter set to 0 and the maximum token count set to 256. 
% and open-source LLaVA's chat instruct interface.
% GPT has the best performance compared to the rest, which is predictable given its higher number of parameters.

% Table \ref{table-baselines} shows the mAP for these approaches, and as you can see, GPT-3.5 has the best performance compared to the rest, which is predictable given its 175B parameters.
% \input{tables/narllm} 

\textbf{Fine-tuned Narrative Captions (NarraCapXL)}
\label{sec:NarracapXL}
Using a small set of human captions obtained over EMOTIC dataset \cite{yang2023contextual} and corresponding ground truth labels, we fine-tuned LLaVA~\cite{liu2023visual} language and vision model to generate captions, and then use these captions and the same prompt used for NarraCaps (see Section~\ref{sec:inference}) to generate top emotions, either directly in LLaVA or by using GPT-4~\cite{openai2023gpt4}.
For finetuning LLaVA, answer format \textit{$<$caption$>$ + This person is feeling $<$labels$>$}, resulted in an F1 score of $30.02$. After changing it to \textit{$<$caption$>$ + $<$name of the target\> is feeling $<$labels$>$}, the F1 score increased to $31.05$. 
LLAVA fine-tuning was performed on
four A40 48GB GPUs. 

\subsection{Vision-Language Model (VLM) Baselines}
We further evaluate three baseline models using Vision-Language Models (i.e., LLaVA and GPT4-vision) to compare the accuracy of these models without the reasoning guidance from the proposed captioning mechanisms. The reasoning capabilities of LLMs and VLMs has been a topic of controversy, where many experiments point towards the lack of common-sense or causal reasoning capabilities in these models \cite{huang2022towards}. Therefore, we cannot claim whether VLMs already should be considered as a Fast \& Slow process, and evaluate these models as a separate baseline category.
We use \textbf{LLaVA} Large Language and Vision Assistant (LLaVA)~\cite{liu2023visual} and Open-AI's \textbf{GPT4-Vision} model as two baselines on large visual-language models. LLaVA is a multi-purpose multimodal model designed by combining CLIP's visual encoder~\cite{radford2021learning} and LLAMA's language decoder~\cite{touvron2023llama}. The model is fine-tuned end-to-end on the language-image instruction-following data generated using GPT-4~\cite{openai2023gpt4}. We used the same prompt in Section~\ref{sec:inference} without the NarraCaps caption. The experiments were run on a single A40 48GB GPU. 
% Include here the fine-tuned version 

For \textbf{Fine-tuned LLaVA} without language guidance, we employ the same dataset as NarraCapXL but without captions. Additionally, we apply the augmentation approach outlined in Sec.~\ref{human-generated-cap}. This experiment was conducted using a single A40 48GB GPU.

\section{Results and Discussion}
The results for our experiments are shown in Table \ref{table-baselines}, and example images with captions in Fig.~\ref{tab:images}.

%\textbf{How do different strategies affect the results?}

Our results showed that, overall, the models that utilized the ``Fast \& Slow" Framework achieved the highest score within their respective trained and non-trained categories, compared to ``Fast"-only or Baseline models.

Among the trained (i.e., EMOTIC) and fine-tuned (e.g., NarraCapsXL, LLaVa) models, we found that the ``Fast\&Slow" Framework performed the best. Guiding the reasoning capabilities with NarraCapsXL captions on both LLaVA and GPT4 proved to be the most successful method. Fast \& Slow framework using NarraCapsXL also surpassed the VLM methods that were not using any reasoning guidance, including the fine-tuned LLaVa, suggesting that VLMs alone may not be enough for emotional reasoning. Apart from achieving the highest F1 score with a small number of fine-tuning examples, the NarraCapsXL method also is very powerful due to its explainability. 

Among the zero-shot methods, as shown in Table \ref{table-baselines}, our results showed that the methods combining ``Fast and Slow" frameworks (NarraCaps+GPT4, ExpNet+GPT4) outperform the methods relying solely on the ``Fast" framework (CLIP) or VLM methods (i.e., LLaVA,GPT4-vision). For instance, in Fig.~\ref{tab:images}.1, it appears that CLIP catgorizes the person with an open mouth as `Surprise' 'Doubt/Confusion' and `Embarrassment'. On the other hand, the `Fast and Slow' frameworks, given the context of a ceromony, or giving a toast, may reason that this body pose is more indicative of `engagement' and `confidence'. 

\textbf{Improving Narrative Captions.}
Within the ``Fast \& Slow" Framework models, we found that ExpansionNet (ExpNet+GPT4) performed the worst, pointing towards the need of providing emotionally relevant captions while guiding the emotion recognition models. However, the zero-shot NarraCaps method was also not as successful as its fine-tuned counterpart (NarraCapsXL), suggesting that the captions themselves need improvement. Upon closer inspection, we found instances where NarraCaps was not as successful in detecting certain actions (how), or characteristics of the humans (who) in the picture. Fig.~\ref{tab:fail} shows example images where the gender and age (3.a), the location (2.a, 4.a) and the action (1.a, 4.a) was detected incorrectly by NarraCaps, while the fine-tuned NarraCapsXL was successful in correctly detecting them. 

To further examine the degree in which the quality of the captions can affect the results, we also directly use the human-generated captions\footnote{https://github.com/Tandon-A/emotic/} were to guide the reasoning of GPT-4. This resulted in an F1 score of 34.17 over the small sample of 387 images~\cite{yang2023contextual}, compared to 26.50 F1 score in the EMOTIC and an 26.19 F1 score with NarraCaps+GPT4. 
These results points toward the promise of guiding emotional reasoning via providing descriptions of the image that focuses on emotionally-salient content. However, NarraCaps method does not reach to human-level performance. We cannot compare these results with NarraCapsXL as it was fine-tuned on this same set.

\begin{figure*}[ht!]
%\vspace{-15mm}
\centering
\includegraphics[width=0.98\linewidth]{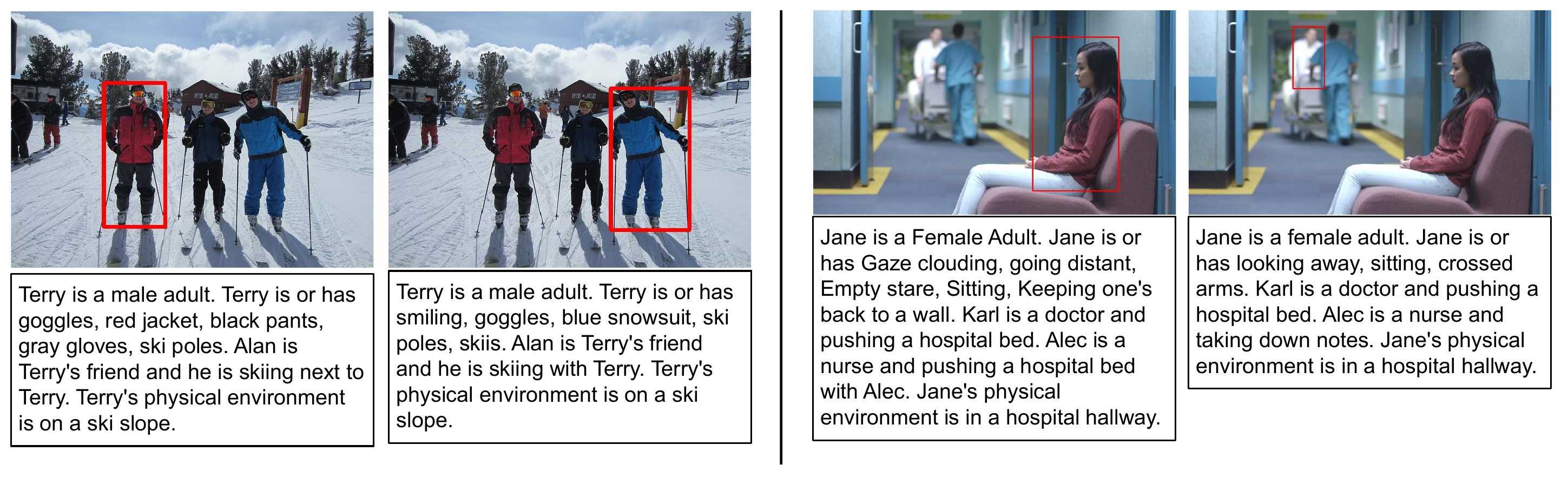}
%\framebox(500,300){}
%\framebox[\textwidth]{\rule{0pt}{140pt}}
\vspace{-2mm}
 \caption{Examples for correct (Left) and incorrect (Right) grounding of NarraCapsXL, where captions change depending on the red bounding box.}
\label{tab:grounding}
\vspace{-4mm}
\end{figure*}
% }]

\begin{figure}[t]
% \vspace{-15mm}
\centering
\includegraphics[width=0.98\linewidth]{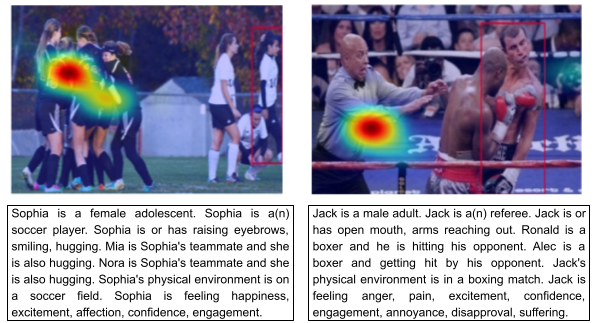}
%{imgs/IROS-CLIP.pdf}
%\framebox(500,300){}
%\framebox[\textwidth]{\rule{0pt}{140pt}}
%\vspace{-2mm}
 \caption{CLIP saliency maps of example images and their resulting captions generated by trained LLaVA model. Captions fail to capture person in the bounding box, even though it explains the scene correctly.}
    \label{tab:clip-map}
\vspace{-4mm}
\end{figure}
% }]
\textbf{The Grounding Problem.}
One of the important shortcomings of the guided reasoning method has been the inability of providing the bounding boxes as input to the system. Previous work showed that CLIP, which we use for generating the one-shot NarraCaps captions and is utilized by multiple VLMs, is unable to capture bounding box information \cite{zhong2022regionclip}. This problem also examined in previous work that showed the inability of VLMs to understand the bounding boxes \cite{wan2024contrastive}. This grounding problem seems to be the biggest challenge in our currently proposed ``Fast \& Slow" Framework, compared to the baseline model where the bounding box information could be provided during training. 

%In the generated captions in Fig.~\ref{tab:grounding} and saliency maps generated by CLIP Fig.~\ref{tab:clip-map}, we see cases where the models fail to focus on the person in the bounding box while generating captions. 

In NarraCaps, we try to overcome this issue by using only the cropped image in the bounding box while asking the ``Who" and ``How" questions, and using the whole image while capturing the environment and actions. However, NarraCapsXL method generated captions using the whole image once. % show examples..
Examples of this grounding problem can be seen in Fig.~\ref{tab:grounding}, where the individuals in each image can have different emotions depending on context and social relationship. 
The saliency map in Fig.~\ref{tab:clip-map} shows that in some cases, where the bounding box is not coinciding with the most salient information, the narration is describing the wrong individual in the image (Fig.~\ref{tab:clip-map}), and resulting in incorrect predictions. This is the case even when the actions and interactions between all individuals in the scene is actually correctly predicted Fig.~\ref{tab:clip-map} (right). The resulting emotion categories can be incorrect in these types of examples. In images where multiple individuals have similar emotions, e.g., Fig.~\ref{tab:grounding} (left), this issue does not result in wrong emotion categorization. 

For our work, we tried to utilize prompt engineering and including the phrase ``Person in the red bounding box is feeling", ``This person is feeling" or ``$<$Name$>$ is feeling", the results of which described in Section~\ref{sec:NarracapXL}. % add F1 scores if we have them

At the time of writing, we are only aware of one method that incorporated the bounding box information to the VLMs \cite{zhang2023llava}. Future work will focus on using one such method while fine-tuning over human captions.

\section{Limitations}
The present study assessing emotional theory of mind on the EMOTIC image dataset using the notion of combining fast visual descriptors with slow reasoning processes. Albeit achieving improvements from the baseline EMOTIC model and being first of its kind, our method is not without limitations. Firstly, we only focused on assessing a few number of Large Language and Vision Language models to test the proposed ``Fast \& Slow" framework capabilities. These included the varying model families offered by OpenAI, including a zero-shot classifier (CLIP), a large language model (GPT), and a large vision language model (GPT4-vision). As GPT-4's vision capability is closed source and has limited access, we use LLaVA as the state-of-the-art open-source alternative for the fine-tuning task. As our paper focuses on examination of guiding the emotional reasoning task inspired by human information processing, we only chose the models that were considered as the top performers at the time of writing and did not focus on a wider breadth of other models for comparison.
%We therefore focus on the breadth of approaches in large language model types that could perform this task, rather than comparing between competing open-source or closed source models within each family (e.g. GPT-3.5 vs. LLAMA). 
It is possible that other models in the family of VLMs and LLM could outperform the investigated models.

%Moreover, we used only ExpansionNet, as an alternative to the both the zero-shot and fine-tuned versions of the proposed Narrative Caption generation. 

In addition, this study focused on studying the as-is capabilities of these models, without a full fine-tuning or training effort, given the EMOTIC dataset. The fine-tuned NarraCapsXL model only used ~300 human-generated captions, most of which were only negative examples. Albeit achieving better accuracy than the baseline, future work needs to include more examples to fine-tune for a better comparison with the baseline models. %There is one exception: we informed each model to provide us top labels based on the average number of labels in the validation set, but did so consistently across all experiments. 
Finally, we did not test our results in other non-contextual emotion recognition datasets in which the environment information or interaction information could not be gathered. However, EMOTIC dataset includes some examples with no visible environment.
%Finally, we noticed that the EMOTIC dataset, while being one of the most challenging image emotion datasets including context, also has small imperfections, including some bounding boxes that contain 2 people instead of only one. Although the case described is rare, a future study could evaluate on other datasets, e.g. one person emotion expression datasets without context, which is considered to be a simpler task.

%\subsection{Limitations of NarraCaps + LLM method}
Lastly, in attempting to guide LLMs and VLMs reasoning capability for this task, we used two methods to generate narrative captions for each person-image pair. One limitation is that we did not perform any quantitative human evalation on the quality of the resulting captions themselves, rather focusing on end-to-end performance against the original human annotated baseline and comparing captions qualitatively. %NarraCaps also did not describe the social interactions or interactions with objects, which if added may increase performance as we see in the LLaVA generated annotations, which were trained on human annotations that included social interactions. 
Future work will focus on quantitative and qualitative examinations of inclusion of the interaction information.

%Moreover, both in NarraCaps usin CLIP and NarraCapsXL versions, we could not ensure that captions were generated for the person in the bounding box. This grounding problem seems to be the biggest challenge in our currently proposed "Fast \& Slow" Framework, compared to the baseline model where the bounding box information could be provided during training. However, as discussed in the results section, future work will be focusing on improving this area, using new methods introduced by researchers \cite{}.
%for the activity and environment detection using CLIP, we performed a standard evaluation with a limited set of classes without a "don't know" or null class, resulting in some mis-captioning. For instance, as shown in Fig.~\ref{tab:fail}.4, the model misunderstood the action as "pillow fight", whereas the action should be described as "covering ears with a pillow", or "yelling". This is seen to be corrected in the fine-tuned captions of NarraCapsXL.  

\section{Conclusion}
In this paper, we explore the capabilities of VLMs and LLMs for the visual emotional theory of mind task by guiding the emotional reasoning capabilities of these models using Narrative Captions. We found that "Fast and Slow" methods (vision and language) outperformed the "Fast" only (vision) frameworks. However, the zero-shot "Fast \& Slow" methods still do not perform better than the models trained specifically for this task on the EMOTIC dataset. The grounding problem seems to be the biggest issue in our method using fine-tuning VLMs. Future work is needed in solving this issue before understanding the full capabilities and shortcomings of the proposed "Fast \& Slow" Framework. However, our results are promising in both improving the performance and providing explainable captions for the emotional recognition. 
% We first find that the zero-shot results were inferior to EMOTIC and Emoticon baseline models which were trained specifically for this task.  
% \yasaman{remove: We observed some potential of the specially captioning for emotion, along with LLM reasoning, as it outperformed the self-supervised vision language models CLIP and LLAVA.} 
%Further research could explore improving the narrative captions by adding other contextual factors to the caption, such as human-object interactions and relationships with other people in the image. Further studies on the characteristics of emotionally comprehensive captioning, coupled with LLMs or with human evaluators could be done.

% \section{Discussion}
% \input{Sections/discussion}

\section{Ethical Considerations}
The research team conducted all the codings and experiments in this study.
The images were sourced from the EMOTIC Dataset, which consists images obtained from internet and public datasets like MSCOCO and Ade20k. To access the EMOTIC dataset, permission must be obtained from the owners. Research team has no affiliation with the individuals depicted in the images. The images show individuals experiencing a range of emotions, such as grief at a funeral, participation in protests, or being in war-related scenarios. However, the dataset does not contain images that surpass the explicitness or intensity typically encountered in news media. 

It is crucial to note that there might be biases affecting this work, such as cultural biases from annotators and biases caused by the languages that GPT and CLIP are trained on. %Furthermore, The source Emotion Thesaurus book is written by North American authors, and similar writing guides in other languages may lead to differing results.
Another limitation of this method is the English language and culture based thesaurus of social signals, written by an North American author. Non-verbal social signals are culture-dependent~\cite{tanaka2010feel} and translation may not suffice. An interesting direction for future work could investigate the distribution of social signals across emotions and between cultures.

The implementation of this work utilized pre-trained models, like GPT, CLIP, and LLaVA. Although no additional training was conducted in this study, it is crucial to acknowledge the significant carbon cost associated with training such models, which should not be underestimated.

As a final note, it is worth mentioning that the authors strongly oppose the utilization of any software that employs emotion estimation approaches with the potential to violate privacy, personal autonomy, and human rights.

\textbf{Examining possible age or gender bias.} 
% One of the advantages of the NarraCaps approach is that it provides a way to explicitly select image details to include for inference (e.g. facial expressions, body pose, relationships) and perform ablations using the text representation. 
One of the advantages of the NarraCaps approach is that it provides a way to explicitly select image details to include for inference and perform ablations using the text representation. 
%To assess the effect of gender, instead of using specific gender labels such as "a baby boy," "a baby girl," "a girl," "a boy," "a woman," "a man," "an elderly man," and "an elderly woman," we only utilized the labels "a female" and "a male." Furthermore, to examine gender, we modified the label list to "a baby," "a kid," "an adult," and "an elderly person."
To investigate the impact of age and gender, we excluded those components from the captions. The results of each study on a set of 1000 images. Our results indicated that including \textit{gender} and \textit{age} in the captions resulted in a slight reduction in F1 scores (a difference of -0.42 and -1.26, respectively), although based on calculations of standard error (SE 0.78 and 0.73, respectively), the difference may be attributed to noise. As a result of this study we initially did not use gender in our experiments, however, to be able to compare with NarraCapsXL that was fine-tuned on human generated captions that included age and gender information, we kept this information in our generated captions. Overall, we found that the impact of gender in our results are negligible and could be omitted from Narrative Captions to guide emotional reasoning. However, age might be a slightly important factor, possibly due to the type of interactions and vulnerability of certain age groups. More evaluations are required to see the extend of which this might be important in emotional reasoning tasks. 

Moreover, we acknowledge that the captions make a statistical categorization of the apparent gender of the individuals and should not be used to infer or make decisions using this information.

%\bibliography{anthology,custom}

\section*{Acknowledgment}
Many thanks to Professor Anoop Sarkar for his invaluable feedback.
\bibliographystyle{IEEEtran} 
%\bibliography{custom}
%\bibliographystyle{plain}
\bibliography{custom}

\end{document}